\documentclass{article}
\usepackage{spconf,amsmath,graphicx}
\usepackage[export]{adjustbox}
\usepackage[symbol]{footmisc}
\usepackage{marvosym}

\title{VISUAL QUESTION ANSWERING INSTRUCTION: UNLOCKING MULTIMODAL LARGE LANGUAGE MODEL TO DOMAIN-SPECIFIC VISUAL MULTITASKS}
\name{Jusung Lee, Sungguk Cha, Younghyun Lee and Cheoljong Yang\textsuperscript{\Letter}\thanks{\Letter{} Corresponding author.}}
\address{NC Research, Republic of Korea}
\begin{document}
%\ninept
%
\maketitle
\begin{abstract}
Having revolutionized natural language processing (NLP) applications, large language models (LLMs) are expanding into the realm of multimodal inputs. Owing to their ability to interpret images, multimodal LLMs (MLLMs) have been primarily used for vision-language tasks. Currently, MLLMs have not yet been extended for domain-specific visual tasks, which require a more explicit understanding of visual information. We developed a method to transform domain-specific visual and vision-language datasets into a unified question answering format called Visual Question Answering Instruction (VQA-IN), thereby extending MLLM to domain-specific tasks. The VQA-IN was applied to train multiple MLLM architectures using smaller versions of LLMs (sLLMs). The experimental results indicated that the proposed method achieved a high score metric on domain-specific visual tasks while also maintaining its performance on vision-language tasks in a multitask manner.

\end{abstract}
\begin{figure}[th]
    \centering
    \includegraphics[width=\columnwidth]{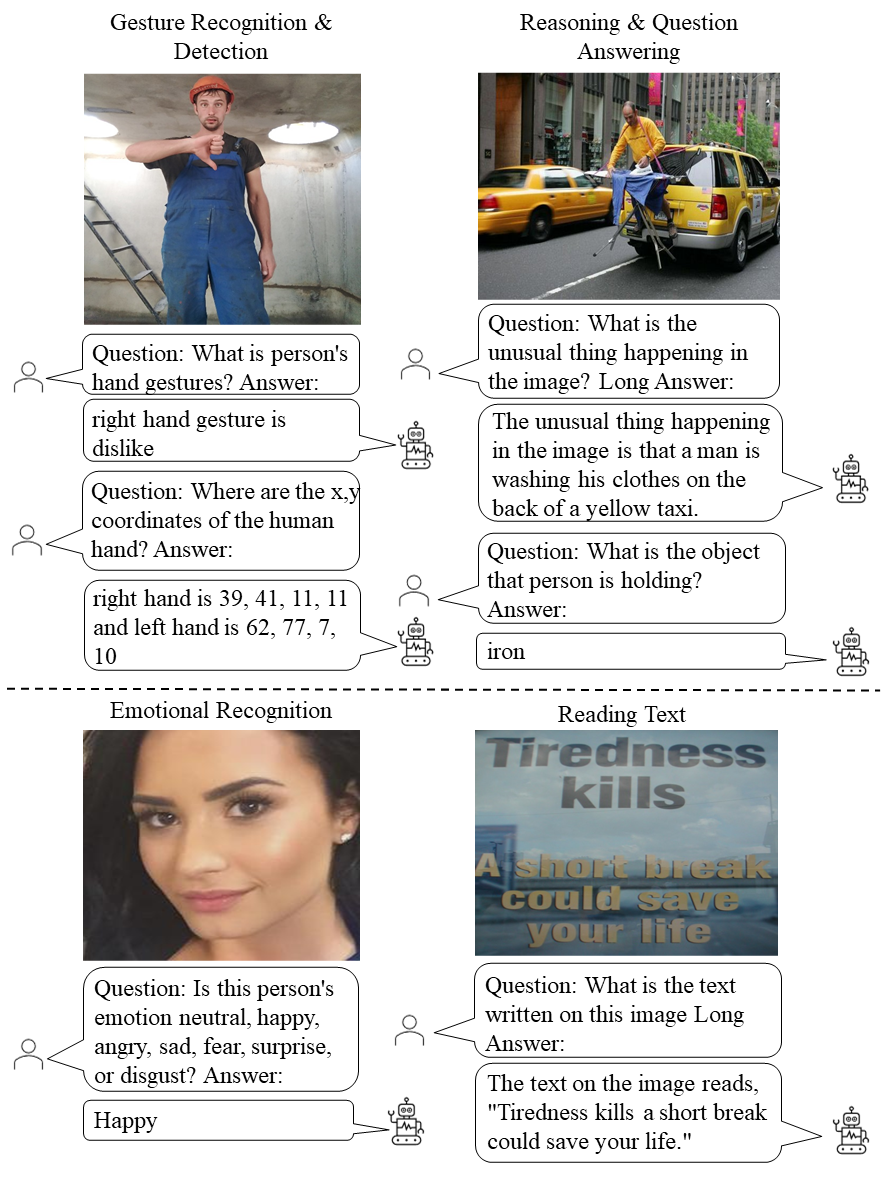}
    \vspace{-1cm}
    \caption{Examples of the proposed visual question answering instruction (VQA-IN), which unlocks MLLMs to enable the creation of a domain-specific visual multitask model.}
    \label{fig:problem_and_motivation}
\end{figure}

\section{Introduction}
\label{sec:intro}
Large language models (LLMs) have significantly improved natural language processing (NLP), exhibiting exceptional performance across a diverse spectrum of language-centered applications. Inspired by the extraordinary performance of conventional LLMs, researchers have designed multimodal Large Language Models (MLLMs) \cite{li2023blip, Dai2023InstructBLIPTG, alayrac2022flamingo} to linguistically interpret visual input.

MLLMs have exhibited a high proficiency in generating textual responses based on images, primarily focusing on vision-language tasks such as image captioning, reasoning or question answering. In constrast, investigations into harnessing MLLMs for traditional vision domain-specific tasks such as recognition or detection have garnered relatively little focus. Our objective was to address this challenge using a visual instruction method, wherein intended questions are aligned with MLLMs in a few-shot or fine-tuned manner.

\begin{figure*}[t]
    \centering
    \includegraphics[width=\textwidth]{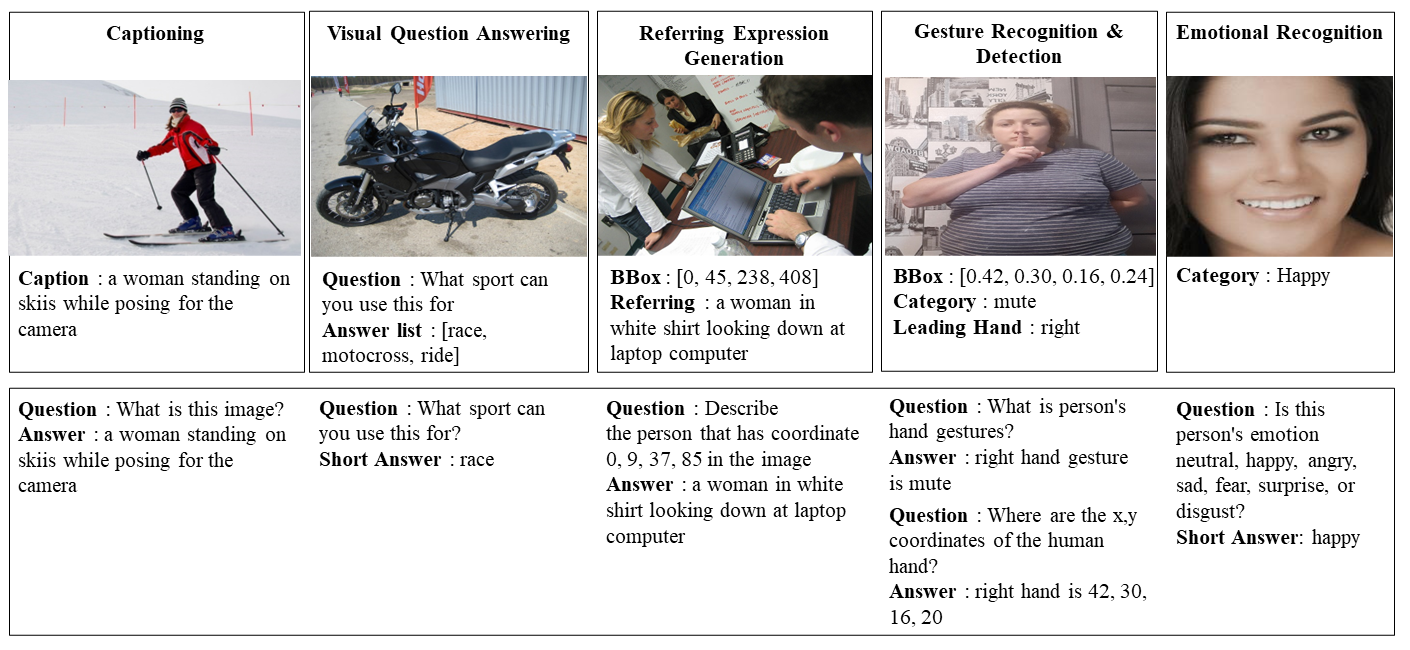}
    \vspace{-0.9cm}
    \caption{Examples of VQA-IN transformation. The first row displays the original dataset format for each task, while the second row shows the results of transforming these diverse formats into a unified question answering format.}

    \label{fig:method}
\end{figure*}

In this study, we developed a method that converts domain-specific visual datasets into the visual question answering instruction (VQA-IN) format. Using this method, the capabilities of LLMs for domain-specific visual tasks can be assessed and multitasking can be performed simultaneously, as shown in Fig. 1. The contributions of this study are summarized as follows.
\begin{itemize}
    \item The VQA-IN  not only transforms vision-language datasets, but also  domain-specific visual dataset, into a unified question answering format. Extensive experiments have been conducted to demonstrate the suitability of the VQA-IN dataset for MLLMs.
    \item This study introduces a visual instruction approach for integrating domain-specific vision tasks into the realm of MLLMs. It offers insights on how MLLMs can effectively utilize instructions to tackle diverse vision tasks.
    \vspace{-0.3cm}

    \item Using VQA-IN transformation, smaller versions of LLMs (sLLMs) could achieve efficient performance in vision-language as well as domain-specific visual tasks within multitask manner.
\end{itemize}

\vspace{-0.4cm}
\section{Related Work}

\label{sec:format}

\subsection{Multi-modal Large Language Model}
\vspace{-0.2cm}

\label{ssec:subhead}
MLLMs typically employ frozen LLMs owing to the scarcity of image-text paired datasets in comparison to text-only datasets, along with the associated training costs \cite{li2023blip, Dai2023InstructBLIPTG, alayrac2022flamingo}. Additionally, Those MLLMs include trainable intermediate layers that merge vision and language embeddings. Flamingo \cite{alayrac2022flamingo} uses cross-attention mechanisms to embed images in LLMs, enabling few-shot learning and the handling of image sequences. On the other hand, BLIP-2 \cite{li2023blip} employs a pretrained Q-former to strike a balance between parameter utilization and performance. The primary objective of MLLM research is to develop a pretrained vision-language model that is predominantly trained using an image-text pair dataset. However, the alignment of these models with user-desired outputs for various vision-language tasks remains challenging due to the scarcity of annotated datasets that provide detailed image information in single sentence.

\vspace{-0.3cm}
\subsection{Instruction Tuning}
\vspace{-0.25cm}
\label{ssec:subhead}
Instruction tuning is a key approach in ensuring that a model aligns with human preferences and effectively completes designated tasks. Although this approach was originally introduced in the NLP domain, it is also applicable to the MLLM domain. The LLAVA model \cite{liu2023visual} demonstrated the effectiveness of incorporating visual instructions to extract detailed explanations from images, whereas InstructBLIP \cite{Dai2023InstructBLIPTG} showed that significant improvements in performance can be achieved in vision-language tasks through the integration of vision instructions using an instruction-aware MLLM architecture. However, existing approaches that incorporate visual instructions use image captions, or are primarily based on the implicit embedding of linguistically rich texts.  Although these methods may be suitable for linguistic tasks, they encounter challenges in extracting visually explicit information.
\begin{table*}[h]
\begin{center}
\caption{\label{tab:Result on VLM Dataset}Vision-language task evaluation results on COCO caption, VQAv2, OKVQA, and GQA. * Due to the absence of evaluation results in previous researches, we use trained model downloaded from LAVIS library \cite{li2022lavis} for evaluation.}
\begin{adjustbox}{width=16cm}
\begin{tabular}{|c|c|c|c|c|c|c|c|c|c|c|}
\hline
\multicolumn{3}{|c|}{Model Info}             & \multicolumn{2}{c|}{Parameters} & \multicolumn{2}{c|}{COCO Caption} & VQAv2 & OKVQA & GQA \\ \hline 
Vision Encoder & LLM              & MLLM Structure                          & Trainable  & Total    &B@4        & CIDEr    & Acc      & Acc      & Acc        \\ \hline \hline
NFNet-F6       & Chinchilla1.4B   & Flamingo3B \cite{alayrac2022flamingo}   & 1.6B       & 3B       & -         & 0.730    & -        & 41.2     & -          \\ 
NFNet-F6       & Chinchilla7B	  & Flamingo9B \cite{alayrac2022flamingo}   & 1.8B       & 9B       & -         & 0.794    & -        & 44.7     & -          \\ 
NFNet-F6       & Chinchilla70B    & Flamingo80B	\cite{alayrac2022flamingo}  & 10B        & 80B	    & -         & 0.843    & -        & 50.6     & -          \\ 
%VIT-L          & OPT2.7B          & BLIP-2 \cite{li2023blip}	            & 104M       & 3.1B     & -         & -        & 50.1     & 30.2     & 30.2       \\ 
%#VIT-G	       & OPT2.7B          & BLIP-2 \cite{li2023blip}	            & 107M       & 3.8B     & -         & -        & 53.5     & 31.7     & 34.6       \\ 
%VIT-G	       & OPT6.7B          & BLIP-2 \cite{li2023blip}	            & 108M       & 7.8B     & -         & -        & 54.3     & 36.4     & 36.4       \\ 
%VIT-L	       & FlanT5XL         & BLIP-2 \cite{li2023blip}	            & 103M       & 3.4B     & -         & -        & 62.6     & 39.4     & 44.4       \\ 
%VIT-G	       & FlanT5XL         & BLIP-2 \cite{li2023blip}	            & 107M       & 4.1B     & -         & -        & 63.1     & 40.7     & 44.2       \\ 
%VIT-G	       & FlanT5XXL        & BLIP-2 \cite{li2023blip}	            & 108M       & 12.1B    & -         & -        & 65.2     & 45.9     & 44.7       \\ \hline \hline
VIT-G	       & OPT2.7B          & BLIP-2* \cite{li2022lavis}	            & 107M       & 3.8B     & 0.329     & 1.110    & 53.5    & 31.8    & 34.5       \\
VIT-G	       & FlanT5XL         & BLIP-2*	\cite{li2022lavis}              & 107M       & 4.1B     & 0.366     & 1.229    & 62.6     & 41.1    & 43.9      \\ 
VIT-G	       & FlanT5XL         & InstructBLIP* \cite{li2022lavis}        & 188M       & 4.2B     & 0.404     & 1.391    & 73.4    & 52.5    & 48.3      \\
VIT-G	       & Vicuna-7B        & InstructBLIP* \cite{li2022lavis}        & 188M       & 7.9B     & 0.410     & 1.407    & 76.6    & 58.1    & 49.1      \\ \hline 
VIT-L	       & OPT1.3B          & OpenFlamingo                            & 386M       & 2.0B	    & 0.274     & 0.930    & 46.5    & 23.4    & 35.3      \\ 
VIT-G	       & OPT1.3B	      & BLIP-2  	                            & 106M	     & 2.4B     & 0.341     & 1.137    & 51.3    & 25.1    & 33.9      \\ \hline 
VIT-L	       & OPT1.3B          & OpenFlamingo                            & 386M       & 2.0B	    & 0.324     & 1.138    & 61.1    & 40.6    & 39.1      \\
               &                  & (VQA-IN)                                &            &          &           &          &         &          &            \\ \hline
VIT-G	       & OPT1.3B	      & BLIP-2  	                            & 106M       & 2.4B     & 0.372     & 1.217    & 54.9    & 30.5    & 36.3      \\ 
               &                  & (VQA-IN)                                &            &          &           &          &         &          &            \\ \hline
VIT-G          & OPT1.3B          & InstructBLIP                            & 188M       & 2.4B     & 0.384     & 1.359    & 70.7    & 45.4    & 43.6      \\ 
               &                  & (VQA-IN)                                &            &          &           &          &          &          &            \\ \hline
%VIT-G          & OPT1.3B          & InstructBLIP & 255M       & 2.5B     & 0.339     & 1.112    & 66.8     & -         & 43.57    & 42.52      \\ 
%               &                  & (VQA-IN)     &            &          &           &          &          &           &          &            \\ \hline
%VIT-G	       & polyglot-ko-1.3b & InstructBLIP & 255M       & 2.5B     & 0.289     & 0.979    & 58.68    & -         & 30.23    & 35.18      \\ 
%               &                  & (VQA-IN)     &            &          &           &          &          &           &          &            \\ \hline
%VIT-G	       & FlanT5XL         & InstructBLIP & 188M       & 4.2B     & 0.378     & 1.245    & 70.48    & -         & 49.90    & 47.55      \\ 
%               &                  & (VQA-IN)     &            &          &           &          &          &           &          &            \\ \hline
VIT-G	       & FlanT5XL         & InstructBLIP                            & 188M       & 4.2B     & 0.380     & 1.249    & 72.6    & 52.3    & 48.6      \\ 
               &                  & (VQA-IN)                                &            &          &           &          &          &          &            \\ \hline
\end{tabular}
\end{adjustbox}
\end{center}
\end{table*}
\vspace{-0.7cm}
\section{Method}
\vspace{-0.2cm}

\label{sec:format}
Numerous datasets of image-text pairs and domain-specific annotations are available for vision-language tasks and domain-specific visual tasks, respectively. However, domain-specific visual datasets might not be directly suitable for MLLMs, as they lack textual data. Furthermore, image-text pair datasets typically contain general textual information pertaining to images, which might not be suitable for domain-specific visual tasks. The VQA-IN method represents a unifying solution that transforms both types of datasets into a consistent format, thereby making them suitable for MLLMs. In Section 3.1 and 3.2, we examine the approaches of VQA-IN by defining two distinct subtasks to account for disparities in annotation styles: vision-language and domain-specific. In Section 3.3, we introduce a method for controlling sentence length by distinguishing between long verbal responses and brief concise answers. Finally, we describe an implementation of MLLM using the proposed VQA-IN.

\vspace{-0.4cm}
\subsection{VQA-IN for Vision-language Task}
\vspace{-0.25cm}
\label{ssec:subhead}
Vision-language datasets typically consist of image-text pairs that can be easily converted into a question answering format, as shown in the caption section of Fig. 2. Regarding to convert texts of caption style to question contents, we utilized a predefined question list and then directly employed question answering pairs. (\textit{i.e. Question: \{random.choice(question\_list)\} Answer:\{captioning\}})

\vspace{-0.4cm}
\subsection{VQA-IN for Domain-specific Visual Task}
\vspace{-0.25cm}
\label{ssec:subhead}
Image recognition is the most common task in domain-specific visual tasks, usually annotated in image-category pair datasets. To utilize these datasets with MLLMs, it is necessary to transform the image-category pairs into a question answering format. To generate appropriate questions, we employed ChatGPT to create suitable question candidates and formulated the answers based on the categories obtained from the corresponding annotations. This process can be exemplified as follows, as shown in the gesture and emotional recognition section of Fig. 2: (\textit{i.e. Question: \{ChatGPT generated question\}  Answer:\{category\}})  

Object detection and reference-expression generation tasks require positional information from images. We represented bounding box coordinates in the form of \textit{\{Bounding Box\}} at the 100th percentile within the text. This approach allows us to avoid architecture changes and the addition of positional tokens, making it easy to integrate into the question answering format as shown in the reference expression generation and gesture detection section of Fig. 2. (\textit{i.e. Question: Describe \{object\} that has coordinate {\{Bounding Box\}} in the image  Answer:\{Reffering\}})

\vspace{-0.45cm}
\subsection{Sentence Length Control}
\vspace{-0.25cm}
\label{ssec:subhead}
One integral aspect of our methodology is the flexibility to adapt sentence lengths using prompts. This mechanism enables the generation of both \textit{\{short\}} or \textit{\{long\}} answers by training question prompts based on predefined word lengths and offering details in responses using prefix prompts. (\textit{i.e. Question:\{Question content\} Short/Long Answer:\{Short/Long Answer content\}})

\vspace{-0.45cm}
\subsection{VQA-IN for MLLM Architecture}
\vspace{-0.25cm}
\label{ssec:subhead}
To demonstrate the effectiveness of our methods across multiple architectures, we trained MLLM architectures from BLIP-2 \cite{li2023blip}, InstructBLIP \cite{Dai2023InstructBLIPTG}, and OpenFlamingo \cite{awadalla2023openflamingo} on VQA-IN datasets. Also, we employed sLLM (parameter size ${<}$ 3B) to demonstrate that our method can be implemented irrespective of state-of-the-art LLMs. 

We pretrained MLLMs with image-text pair datasets as used in \cite{li2023blip} and then finetuned on VQA-IN datasets which is consist of VQAv2 \cite{goyal2017making}, OKVQA \cite{marino2019ok}, LLAVA 150k \cite{liu2023visual}, HaGRID \cite{kapitanov2022hagrid}, AffectNet \cite{mollahosseini2017affectnet} and image-text pair datasets used in pretraining stage. We used a balanced dataset sampling strategy, as mentioned in \cite{Dai2023InstructBLIPTG}, to prevent overfitting of small datasets. The AdamW optimizer \cite{loshchilov2017decoupled} with $\beta_{1} = 0.9$ and $\beta_{2} = 0.999$ is used.  The learning rate starts from $10^{-8}$ and warmed up to  $5\times10^{-5}$ , and then decayed with a cosine schedule to $10^{-8}$. 

\begin{table}[tb]

\begin{center}
\caption{\label{tab:Result on Vision-Centric Dataset}Domain-specific visual task evaluation results on AffectNet, HaGRID, RefCOCOg.}
\begin{adjustbox}{width=8.5cm}
\begin{tabular}{|c|c|c|c|c|}
\hline
                                              & AffectNet  & HaGRID  & RefCOCOg \\ \hline
Model                                         & Acc        & Acc     & METEOR             \\ \hline \hline
VGG-FACE \cite{kollias2020deep}               & 60.00      & -       & -                  \\ \hline
%DAN \cite{wen2023distract}                   & 65.69      & -       & -                  \\ \hline
HaGRID ResNet-152 \cite{kapitanov2022hagrid}  & -          & 94.49   & -                  \\  
HaGRID ResNeXt-101 \cite{kapitanov2022hagrid} & -          & 95.65   & -                  \\ \hline
MMI \cite{mao2016generation}                  & -          & -       & 0.144              \\  
visdif \cite{yu2016modeling}                  & -          & -       & 0.151              \\ 
KOSMOS-2 \cite{peng2023kosmos}                & -          & -       & 0.122              \\ 
attr \cite{liu2017referring}                  & -          & -       & 0.163	          \\ \hline \hline
BLIP-2${_{OPT1.3B}}$                           & 5.02       & 25.60   & 0.093              \\  
BLIP-2${_{FlanT5XL}}$                          & 14.43      & 17.78   & 0.061              \\  
InstructBLIP${_{FlanT5XL}}$                    & 21.20      & 20.96   & 0.043              \\  
InstructBLIP${_{Vicuna-7B}}$                   & 28.00      & 5.68    & 0.029              \\ \hline \hline
BLIP-2${_{OPT1.3B}}$(VQA-IN)                     & 54.60      & 93.77   & 0.163              \\  
OpenFlamingo${_{OPT1.3B}}$(VQA-IN)            & 52.31      & 94.72   & 0.192              \\ 
InstructBLIP${_{OPT1.3B}}$(VQA-IN)            & 58.49      & 97.49   & 0.225              \\  
InstructBLIP${_{FlanT5XL}}$(VQA-IN)            & 57.77      & 96.22   & 0.187              \\ \hline 

\end{tabular}
\end{adjustbox}
\end{center}
\vspace{-0.4cm}

\end{table}

\vspace{-0.4cm}

\section{Experiments}
\label{sec:pagestyle}
\vspace{-0.1cm}

We conducted two evaluation phases with respect to task type – vision-language tasks and domain-specific visual tasks – as the former are well-established and benchmarked in MLLM research whereas the latter are not.
\vspace{-0.4cm}
\subsection{Evaluation on Vision-language Tasks}
\vspace{-0.2cm}

\label{ssec:subhead}
{\bf Datasets}
The evaluation of vision-language capabilities was performed on the COCO Captions \cite{lin2014microsoft}, VQAv2 \cite{goyal2017making}, OKVQA \cite{marino2019ok}, and GQA \cite{hudson2019gqa} datasets, which have commonly been used in previous studies \cite{li2023blip}, \cite{alayrac2022flamingo}. We present evaluation results for the validation set in VQAv2 \cite{goyal2017making}. In vision-language tasks, we present the BLEU and CIDEr score for COCO Captions. For the other datasets, we report VQA accuracy. It is worth noting that higher values indicate better performance in all metrics.

{\bf Experiment Results}
As shown in Table 1, MLLMs trained with VQA-IN outperformed the baseline BLIP-2 \cite{li2023blip} and OpenFlamingo \cite{awadalla2023openflamingo} model in the vision-language task. OpenFlamingo${_{OPT1.3B}}$(VQA-IN) achieved equivalent performance to Flamingo3B with two-thirds of the parameter size. BLIP-2${_{OPT1.3B}}$(VQA-IN) achieved an equivalent performance to BLIP-2${_{OPT2.7B}}$ with less parameters.  InstructBLIP${_{FlanT5XL}}$(VQA-IN) achieved a comparable performance with InstructBLIP${_{FlanT5XL}}$,  despite the latter being heavily instructed with several VQA datasets. InstructBLIP${_{OPT1.3B}}$(VQA-IN) exhibited minimal performance degradation with an almost twofold decrease in parameters compared to InstructBLIP${_{FlanT5XL}}$. Notably, instructions contributed to enhancements in general vision-language capabilities, as evidenced by performance on the GQA dataset \cite{hudson2019gqa}, wherein instructions were not utilized for VQA-IN.
\vspace{-0.3cm}
\subsection{Evaluation on Domain-specific Visual Tasks}

{\bf Datasets}
To assess domain-specific visual capabilities, we chose AffectNet \cite{mollahosseini2017affectnet}, HaGRID \cite{kapitanov2022hagrid} and RefCOCOg  \cite{mao2016generation} dataset. AffectNet \cite{mollahosseini2017affectnet}, an emotion recognition dataset comprising face images cropped to focus on the facial region, was used for the domain-specific recognition task. HaGRID \cite{kapitanov2022hagrid}, a hand gesture recognition image dataset specifically compiled to provide videoconferencing services catering to people who are speech- or hearing-impaired, was used for the domain-specific recognition and detection tasks. RefCOCOg \cite{mao2016generation} is a reference expression dataset used to train models to interpret positional information and distinguish individual objects based on images. We used RefCOCOg \cite{mao2016generation} dataset to evaluate the referring expression generation task. Furthermore, we used the validation sets of AffectNet \cite{mollahosseini2017affectnet} and RefCOCOg \cite{mao2016generation}, as well as the test sets of HaGRID \cite{kapitanov2022hagrid}, to evaluate model performance. For quantitative evaluation, we present the top-1 accuracy for the recognition task, and we report the METEOR metric \cite{denkowski2014meteor} for the referring task.

{\bf Prompts}
To ensure consistent answer formats in the VQA-IN setup, we applied minor modifications to questions for comparative purposes. Specifically, we employed the following prompts: \textit{"Question: What is the person's emotion in \{Category\}? Short Answer:"} for AffectNet \cite{mollahosseini2017affectnet}, \textit{"Question: What is the person's hand gesture in \{Category\}? Answer:"} for HaGRID, and \textit{"Question: Describe the person with coordinates \{Bounding Box\} in the image. Answer:"} for RefCOCOg \cite{mao2016generation}.

{\bf Experiment Results}
To evaluate MLLMs trained with VQA-IN on domain-specific visual tasks, we compared our results with those of domain-specific models and MLLMs. InstructBLIP${_{OPT1.3B}}$(VQA-IN) achieved best performance on domain-specific visual tasks, while InstructBLIP${_{FlanT5XL}}$, OpenFlamingo${_{OPT1.3B}}$, and BLIP-2${_{OPT1.3B}}$ trained with VQA-IN exhibited significant improvements compared to InstructBLIP${_{FlanT5XL}}$, InstructBLIP${_{vicuna-7B}}$, and BLIP-2${_{OPT1.3B}}$, as shown in Table 2. The baseline MLLMs exhibited weaker performance than that of the single domain-specific model. This discrepancy may be attributed to the image-text pair dataset, which does not comprehensively cover the range of domain-specific tasks. However, VQA-IN addressed this discrepancy by converting domain-specific data into a question answering format, thereby achieving nearly equivalent performance to that of each domain-specific model within a single multitask model. Notably, even small LLMs have demonstrated improvements when applied to VQA-IN, leading to a consistent performance regardless of the type of LLM or parameter size. 

\vspace{-0.2cm}
\section{Conclusions}
\vspace{-0.1cm}

\label{sec:typestyle}

VQA-IN extended the capabilities of multimodal LLMs to domain-specific visual tasks and effectively managed these tasks within a multitask framework without compromising vision-language capabilities. The results of this study help the practical application of MLLMs in the development of multimodal assistants, enabling them to handle a diverse range of visual tasks.
\vfill\pagebreak


\bibliographystyle{IEEEbib}
\bibliography{strings,refs}

\end{document}